\documentclass[sigconf,nonacm]{acmart}

\usepackage{booktabs}
\usepackage{amsmath}
\usepackage{csquotes}
\usepackage{tikz}
\usetikzlibrary{arrows.meta, positioning, calc, shapes.geometric}

\AtBeginDocument{%
  }

\begin{document}

\title{Event Cameras versus Frame Cameras for Synthetic-Aperture Imaging: A Simulation Study}

\author{Jakob Hinterholzer}
\affiliation{%
  \institution{Johannes Kepler University Linz}
  \department{Institute of Computer Graphics}
  \city{Linz}
  \country{Austria}}
\email{jakob\_hinterholzer@outlook.de}

\author{Mohamed Youssef}
\affiliation{%
  \institution{Johannes Kepler University Linz}
  \department{Institute of Computer Graphics}
  \city{Linz}
  \country{Austria}}
\email{mohamed.youssef@jku.at}

\author{Oliver Bimber}
\affiliation{%
  \institution{Johannes Kepler University Linz}
  \department{Institute of Computer Graphics}
  \city{Linz}
  \country{Austria}}
\email{oliver.bimber@jku.at}

\renewcommand{\shortauthors}{Hinterholzer, Youssef and Bimber}

\begin{abstract}
Synthetic-aperture imaging (SAI) removes occluders such as foliage by integrating many views of a scene along a camera path. It has so far been used mainly with frame cameras. We ask whether an event camera has advantages for SAI, and answer the question in simulation, where both sensors see exactly the same scene and the same camera movement. A densely rendered Blender sequence is turned into frames with an EMVA~1288 noise model of an industrial global-shutter camera, and into an event stream with the v2e simulator parameterized like a Sony IMX636. The frames are integrated with the existing airborne optical sectioning renderer. For the events we add a new integration path that projects every single event onto the focal plane at its own interpolated pose, so no events have to be binned into frames. We compare the two sensors in three main runs and an additional dark high-speed test. In a bright scene at $25$\,m/s the frame camera records only three images on the path and cannot remove the occluder, while the event camera reconstructs the target, even after an event-rate controller drops most of its events to fit the camera link. At the same speed in a $5$\,lux scene the event camera no longer resolves the target either. In a $5$\,lux scene at $5$\,m/s both sensors reconstruct the target at similar quality. We also compare the data one traversal of the aperture costs. At high speed the capped event stream still shows the target at the same data volume as the three frames. At $1$\,m/s we compare both cameras at two matched data budgets, and there the event camera has no advantage over the frame camera in reconstruction quality. The data rate of the event camera depends on the scene and is highest exactly in the fast bright scenes where it is needed most.
\end{abstract}

\maketitle

\section{Introduction}

Synthetic-aperture imaging (SAI) is a technique used to see through occlusions, such as leaves on a tree.
It fuses images captured by a moving camera and requires the 3D position or pose of the camera for each image to be known.
With the 3D positions of each frame and a given distance from the camera, it is possible to focus on objects behind the occlusion, while the occluder, which is positioned at a different distance from the camera, is averaged out. Airborne optical sectioning (AOS) has shown this on a drone system that can be used to see objects of interest on the floor of a forest~\cite{kurmi_aos_2018}.
In addition, this has also been used on ground-moving systems that are able to look through bushes or other vertical occluders~\cite{bimber_robot_dogs_2025}.
The technique has been mainly shown on frame-based sensors, including thermal cameras. However, frame cameras are limited by their frame rate. If not enough frames are captured, the occlusion removal does not work.
Exposure must also be managed carefully to avoid motion blur.

Event cameras are often proposed as an alternative sensor that has advantages over classic frame-based sensors~\cite{lichtsteiner_dvs_2008, gallego_survey_2022}.
In contrast to frame-based cameras, event cameras record lighting changes over a certain threshold and therefore have no frame rate. Each pixel reports changes independent of the pixels around it.
Especially in difficult lighting conditions and in scenarios where motion blur is an issue, it has been demonstrated that event cameras perform better~\cite{sun_frames_events_2021, rebecq_e2vid_2021}.
Also, in many scenarios, they have a bandwidth advantage since event cameras only record pixels if something changes, and frame cameras have a static bandwidth~\cite{gallego_survey_2022}.

We study the question of whether event cameras have advantages over frame cameras in the context of synthetic-aperture imaging. We use simulation with a shared trajectory and scene to compare both modalities. Both simulation pipelines start with a dense Blender render and are then fed into sensor models. For the event camera simulation, we use an event camera simulator (v2e~\cite{hu_v2e_2021}).
For the frame sensor simulation, we use an EMVA~1288~\cite{emva1288}-parameterized simulation. The EMVA parameters come from an industrial global-shutter camera. The frame simulation is then forwarded to the existing SAI technique.
The existing SAI takes images, so events could only be used by binning them back into frames. Instead we add an SAI path that projects every single event at its own interpolated pose, so it also works with arbitrary 6-DoF poses. Event-based SAI has so far been shown on a linear rail~\cite{zhang_esai_2021, yu_seethrough_2022}.
The simulation lets us sweep different speed and lighting scenarios and compare the performance of both modalities.

We evaluate three runs. The first two cover the two conditions in which event cameras are expected to help: fast motion and low light. The high-speed scenario is simulated in bright light and at a camera speed of $25$\,m/s. In this scenario, the event camera provides a usable result, while the frame camera, running at its maximum full-resolution frame rate (Section~\ref{sec:setup}), records only three images across the aperture, which is not enough to suppress the occluders.
However, we had to simulate an event rate controller for this result to be valid, since the event simulator produced too many events, exceeding the bandwidth of the link of a real camera. Even with the limited bandwidth, the event camera still provides a usable result. This advantage is limited to good light. In the dark at the same speed the event camera does not resolve the target either (Section~\ref{sec:discussion}). For a drone with a $30$\,fps camera the gain in flight speed is about a factor of $17$ on our aperture.
In the low-light scenario, the camera speed is set much lower, at $5$\,m/s. For this scenario, no decisive winner could be found. This is partly because we simulate a higher-resolution event camera, and such cameras generally do not perform as well in low light as lower-resolution event sensors~\cite{gehrig_high_res_2022}. The frame camera images are quite noisy, but the SAI reconstruction is able to remove most of that noise through averaging.
The third run is in the bright scene at $1$\,m/s, where the frame camera has enough views. There we compare both cameras at matched data budgets, and the event camera shows no advantage.

Our comparison deliberately uses uniform illumination, so the event camera's
high-dynamic-range advantage is excluded by construction, since the event simulator cannot generate HDR scenarios.
Our contributions are:
\begin{itemize}
\item an event-native synthetic-aperture integration: each event,
back-projected to a ray at its individually interpolated
continuous-time pose, is splatted onto the focal plane with an
area-normalized footprint, extending the existing pose-driven AOS
renderer to event streams with arbitrary 6-DoF poses, where prior
event-based SAI assumed a linear trajectory;
\item a simulation pipeline that compares event-based and frame-based SAI
from a single shared trajectory, with both sensor models anchored to
manufacturer data, including a model of the event-rate controller;
\item an evaluation in three runs and one additional dark high-speed run: at high speed only the event camera
reconstructs the target, but only in good light; in low light both sensors reach similar
results; at $1$\,m/s with matched data budgets the event camera has no
advantage over the frame camera, so the bandwidth advantage we see is
limited to very high speed.
\end{itemize}

\section{Related Work}

\paragraph{Synthetic-aperture imaging through occlusion.}
Seeing through occluders by combining many viewpoints goes back to the
Stanford camera array. Vaish et al.~\cite{vaish_calib_2004} showed that
a dense grid of conventional video cameras can be combined and refocused
as if they were one wide-aperture sensor. Wilburn et
al.~\cite{wilburn_arrays_2005} built the full 100-camera system and used
it for synthetic-aperture video, blurring foliage away to focus on a
subject behind it. Airborne optical sectioning (AOS)~\cite{kurmi_aos_2018}
brought the principle to a single low-cost drone, which made it more
practical and easier to use. The drone flies over a point of interest
for about 30 to 100\,m and collects hundreds of video frames. After
that, the camera poses are calculated in post-processing. Later this
principle was also showcased with thermal cameras. Combining many
thermal images revealed heat signatures through trees that no single
image would have shown~\cite{kurmi_thermal_2019}. Another project using this principle was a
fully autonomous drone that was able to locate 38 of 42 hidden persons
in field trials~\cite{schedl_drone_2021}. Most recently,
synthetic-aperture imaging has been transferred to ground
robots~\cite{bimber_robot_dogs_2025}. In all of this work, the sensor is
a frame camera.

\paragraph{Event-based synthetic-aperture imaging.}
Event cameras have also been used for synthetic-aperture imaging.
Zhang et al.~\cite{zhang_esai_2021} mounted a DAVIS346 on a motorized
linear rail and moved it past a wooden fence at a constant speed of
about $0.2$\,m/s. The events are refocused with a single shift that
follows from the camera speed and the target depth, and a hybrid
spiking and convolutional network, trained on their own recordings,
then reconstructs the hidden target from the refocused events. The
frame comparison in this work uses the APS frames of the same
DAVIS346. Liao et al.~\cite{liao_efsai_2022} combined both modalities,
using the frames for the texture that events miss and the events for
the see-through signal under dense occlusion. Yu et
al.~\cite{yu_seethrough_2022} extended the method with a network that
estimates the camera speed and target depth itself and enlarged the
dataset. These all work on a single axis and at table-top scale. Our
implementation, like AOS, works with six degrees of freedom. It is an
event extension of the AOS software and therefore also works without a
trained model.

\paragraph{Frame versus event comparisons.}
Event cameras are an alternative sensor category. Their advantages over
frame cameras have been demonstrated directly in side-by-side
comparisons, though for tasks other than synthetic-aperture imaging.
Delbruck and Lang~\cite{delbruck_goalie_2013} established the
low-latency advantage of events for situations where a fast reaction is
crucial. Sun et al.~\cite{sun_frames_events_2021} mounted an event and a
frame sensor on one quadrotor and compared them across different
scenarios, finding the event camera's advantage concentrated in fast
motion and low light. Gehrig and Scaramuzza~\cite{gehrig_high_res_2022}
showed, however, that these advantages depend on the specific sensor and also on the resolution of the sensor:
high-resolution event cameras produce more noise events in dim scenes
and can saturate their readout bandwidth in busy scenarios. Newer sensors therefore include an
event-rate controller that drops events~\cite{prophesee_gen41}, a mechanism we reproduce in our ERC model.

\paragraph{Event cameras in low light.}
This low-light weakness has a physical origin. Since the original
DVS~\cite{lichtsteiner_dvs_2008}, it is known that the performance of event cameras is strongly influenced by the illumination level they are used in:
an event pixel is driven by the light it
receives, so in scenes with a lot of light it reacts quickly and in lower-light scenes it
gets slow and cannot follow fast brightness changes and its output
gets noisier~\cite{graca_dvspixel_2023, graca_noise_2021}. Smaller
pixels in higher resolution systems collect less light and therefore
slow down earlier as the light decreases~\cite{gehrig_high_res_2022}.
How fast a pixel can still follow brightness changes is expressed by a
single number, its cutoff frequency.
The event generation model we use needs this number to simulate a
certain light condition, but for the current-generation sensor we
model~\cite{finateu_imx636_2020}, no exact measurement of it at low
light exists~\cite{mcreynolds_step_2024}. But the manufacturer does
specify the pixel latency at specific lux
levels~\cite{prophesee_gen41}. We therefore derive the cutoff frequency
from the latency specification and choose a value below both
estimates we have.

\paragraph{Event-camera simulation.}
There are several simulators that turn a video into events. They all
use the same base idea: the video is processed frame by frame, and
then for each pixel the simulator checks if the brightness has changed
by more than a contrast threshold since the last event. If it has, an
event is written out. The main difference between the simulators is how
much of the real pixel they model on top of this.
ESIM~\cite{rebecq_esim_2018} and vid2e~\cite{gehrig_vid2e_2020} use an
ideal pixel that does nothing more than this comparison, so the events
they produce have no noise. v2e~\cite{hu_v2e_2021} goes further and also
models the weaknesses of a real DVS pixel. The photoreceptor bandwidth,
the threshold mismatch, shot and leak noise and the refractory period
are all parameters that can be set. What cannot be set is the light
level of the scene. In v2e a lighting condition is simulated by giving
these parameters the values a real pixel would have under that light.
For low light this mainly means a low photoreceptor bandwidth and more
noise. Our night study is exactly the case where these weaknesses
matter most~\cite{graca_dvspixel_2023}, and therefore we use v2e.

\paragraph{High dynamic range.}
Another event-camera advantage that is reported often is dynamic
range. In comparison to a normal frame camera there is no uniform
exposure for all pixels. Instead each pixel responds logarithmically,
and therefore a much higher dynamic range can be recorded. A DVS
covers 120\,dB within one scene~\cite{lichtsteiner_dvs_2008} without
exposure control~\cite{gallego_survey_2022}, against 72\,dB for the
frame camera we use as reference~\cite{emva1288, basler_emva_overview}.
This has been shown many times for scenes that saturate or starve a
frame camera, such as tunnel exits and headlight
glare~\cite{rebecq_e2vid_2021, sun_frames_events_2021,
liu_nighttime_2024}. This could definitely be an advantage in some
scenarios for SAI, but a multi-image system can
also expose each image differently and combine
them~\cite{wilburn_arrays_2005}. The event simulator we use, however,
does not support HDR, since its input consists of 8-bit frames, and
therefore our simulation does not test this question. But this
advantage should definitely be taken into consideration when deciding
which sensor to use.

\section{Method}
\label{sec:method}
\begin{figure}[t]
\centering
\begin{tikzpicture}[
  font=\footnotesize, >={Stealth[length=1.8mm]},
  box/.style={draw, rounded corners=2pt, align=center, minimum height=8mm,
              inner sep=3pt, text width=32mm},
  wide/.style={box, text width=70mm},
  stage/.style={font=\scriptsize\itshape, gray, anchor=east},
  node distance=4mm and 5mm]
  \node[wide] (render) {Blender rendering\\
    dense samples along one trajectory, pose per sample};
  \node[box, below left=of render.south, anchor=north east, xshift=-1mm]
    (fcam) {Frame-camera model\\ exposure, sensor noise};
  \node[box, below right=of render.south, anchor=north west, xshift=1mm]
    (ecam) {Event-camera model (v2e)\\ low-pass filter, thresholds};
  \node[box, below=of fcam] (fint) {Frame integration\\
    AOS image integration};
  \node[box, anchor=north] at (fint.north -| ecam) (eint)
    {Event integration\\ per-event splatting (new)};
  \node[wide, below=of $(fint.south)!0.5!(eint.south)$, anchor=north]
    (out) {Integral images on the same focal plane};
  \draw[->] (render.south) -- ++(0,-2mm) -| (fcam.north);
  \draw[->] (render.south) -- ++(0,-2mm) -| (ecam.north);
  \draw[->] (fcam) -- (fint);
  \draw[->] (ecam) -- (eint);
  \draw[->] (fint.south) |- ($(out.north)+(0,2mm)$) -- (out.north);
  \draw[->] (eint.south) |- ($(out.north)+(0,2mm)$) -- (out.north);
  \node[stage, left=2mm of render.west] {1};
  \node[stage] at ($(fcam.west)+(-2mm,0)$) {2};
  \node[stage] at ($(fint.west)+(-2mm,0)$) {3};
\end{tikzpicture}
\caption{Overview of the pipeline with its three stages.}
\Description{Flowchart with one box at the top for the Blender rendering,
two parallel columns below it for the frame-camera and event-camera
models and their integrations, and one box at the bottom for the
resulting integral images.}
\label{fig:pipeline}
\end{figure}
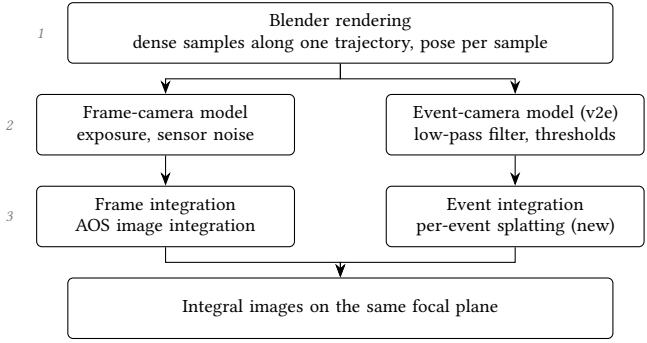

The pipeline has three stages (Figure~\ref{fig:pipeline}). First, one
camera trajectory past the occluded scene is rendered densely in Blender,
together with the camera pose of every sample. Second, this rendering is
fed into two sensor models: a frame-camera model that turns groups of
samples into exposed, noisy images, and an event-camera model that turns
the sample sequence into an event stream. Third, each output is
integrated on the same focal plane. The frames use the existing SAI of
the AOS renderer, and for the events we add a new SAI path that projects
every single event onto the focal plane. The rendering in Blender, the
movement of the camera and the focal plane are shared, so the difference
between the two results comes from the sensor model and its
reconstruction. The following subsections describe each stage.

\subsection{Rendering and Pose Extraction}
\label{sec:render}
The input to both sensor models is a single densely sampled Blender
rendering of one camera trajectory past the occluded scene. The path of
the camera forms the synthetic aperture. The trajectory is built from a
chain of movement commands. Each command translates the camera over a
specified distance, with optional parameters for rotation and elevation
changes. Consequently, a combination of these commands can generate many
different sequences of camera poses. The animation is sampled at $N$
positions and each sample is rendered as an 8-bit grayscale image,
matching the monochrome sensors we model, together with its camera pose.
For every rendered frame the camera pose from Blender, which places the
camera in the world, is inverted into the world-to-camera view matrix
expected by the reconstruction. This requires remapping Blender's
$Z$-up coordinate convention to the renderer's $Y$-up, $Z$-forward one.
All poses are collected in a single JSON file, which is then used by the
SAI reconstruction.

\paragraph{One render, many speeds.}
The rendered positions are pure geometry: no timestamps are attached at
render time. Camera speed is configured by setting a specific duration
$T$ of the movement, giving an effective input rate
\[
  f_{\text{eff}} = N / T .
\]
The same render can therefore be used as a fast and a slow traversal of
the same camera movement, and both sensor branches consume the same
reinterpreted sequence, so the two camera simulations always have the
same input. The movement duration parameter propagates through the
whole pipeline. It sets the inter-frame interval of the event
simulator's input, which must remain small against the photoreceptor
time constant (Section~\ref{sec:eventmodel}). It sets the number of
render frames averaged into one frame-camera exposure. And it sets the
number of frames the frame camera gets from a specific frame rate across
the motion, $n = f_{\text{sensor}} \cdot T$. The number of frames
therefore depends only on $T$, while the number of events depends on the
motion in the image and on the \texttt{cutoff\_hz} we set in v2e. The
render density $N$ is chosen for the fastest movement we want to
simulate, since v2e requires shorter inter-frame intervals for higher
\texttt{cutoff\_hz} settings.

\subsection{Event-Camera Model}
\label{sec:eventmodel}
The event camera simulation uses v2e~\cite{hu_v2e_2021} to convert the
densely rendered frames into the event stream a dynamic vision sensor
(DVS)~\cite{lichtsteiner_dvs_2008} would produce. Frames are read one at
a time, and for each pixel the simulator calculates the following. First,
the log of the brightness value of each pixel is calculated, since event
cameras process brightness relatively. Then this log signal is sent into
a low-pass filter that models the pixel's bandwidth restrictions.
Finally, events are generated from the \emph{filtered} signal wherever
the signal goes over the threshold.

\paragraph{Photoreceptor low-pass and \texttt{cutoff\_hz}.}
When an event camera pixel receives light, a photocurrent is induced,
and the photoreceptor converts this current into a logarithmic voltage.
The time constant of this conversion is inversely proportional to the
photocurrent~\cite{hu_v2e_2021, lichtsteiner_dvs_2008}. With less light
there is less photocurrent, the time constant becomes longer, and the
bandwidth of the pixel drops, so pixels react quickly in bright light
and sluggishly in dim light~\cite{hu_v2e_2021, graca_dvspixel_2023}.
Therefore the photoreceptor can be modeled using a low-pass filter,
which is what v2e does. The event sampling itself is set after the
low-pass filter, and therefore the events are generated on a smoothed
signal of the rendered frames~\cite{hu_v2e_2021}. The parameter
\texttt{cutoff\_hz} sets the corner frequency of this filter, and v2e
scales the filter per pixel with its brightness, so \texttt{cutoff\_hz}
specifies the bandwidth of a bright pixel and darker pixels respond
proportionally slower for the same cutoff setting. If
\texttt{cutoff\_hz} is low enough, finer structures, like the spokes of
a bike, produce fewer or no events. This is the reason why in lower
light the event stream loses detail. \texttt{cutoff\_hz} is not a
parameter of the sensor alone, but a combination of the event camera,
its bias settings and the brightness of the
scene~\cite{graca_dvspixel_2023, hu_v2e_2021}. How we choose its value
per lighting condition is described in Section~\ref{sec:setup}.

\paragraph{Threshold events and non-idealities.}
Each pixel saves the log brightness at which it last fired an event.
When the filtered signal for this pixel differs from this stored value
by more than the threshold, the pixel emits an ON or OFF event and
updates the stored value, one event per threshold crossed. Events within
one frame interval get timestamps between the two frame times. v2e
spaces them on an even grid whose resolution is set by the pixel with
the most events in that frame~\cite{hu_v2e_2021}. On top of this, v2e
adds the non-idealities of a real sensor: every pixel gets a slightly
different threshold, pixels fire random noise events at a rate that
depends on the light, some pixels fire leak events without any input,
and a pixel cannot fire again within a short refractory time.

\subsection{Frame-Camera Model}
\label{sec:framemodel}
The frame camera simulation reuses the same rendered frames as the event simulation. The simulation has two steps. The first step takes the shutter speed and applies motion blur based on it. The second step applies noise based on the shutter speed and the EMVA~1288 camera model~\cite{emva1288}. This is a standardized model and many manufacturers publish parameters for it~\cite{emva1288, basler_emva_overview}.

\paragraph{Exposure.}
The rendered frames are sRGB encoded, so we first convert them back to linear brightness. Only in linear space can light from several frames be added up correctly. Each camera frame is then the average of all render frames that lie inside its exposure time. This is the common way to get motion blur from densely rendered sharp frames~\cite{nah_deblur_2017}. The camera frame gets the pose of the render frame in the middle of its exposure. The exposure cannot be longer than the time between two frames, since the sensor has to read out before the next frame starts. So for darker scenes the capped exposure results in more noise. In SAI noise is generally preferred to motion blur, since the motion blur cannot be averaged out.

\paragraph{Sensor noise.}
For the sensor stage we use the EMVA~1288 camera model~\cite{emva1288}. It describes how much light a pixel collects during the exposure and how noisy the recorded value is. The main noise source is shot noise, which comes from the random arrival of photons and gets worse the less light a pixel collects. On top of that, the readout adds a constant read noise. Very bright pixels saturate, and the result is written as an 8-bit image. The model needs only four numbers from the camera: quantum efficiency, pixel pitch, read noise and full-well capacity. We take all four from the EMVA~1288 measurement the manufacturer publishes for the camera we simulate (Section~\ref{sec:setup}). The exposure is set so that a white surface fills the output range, like an auto exposure would do.

\subsection{Synthetic-Aperture Reconstruction}
\label{sec:sai}
Both sensor simulations end on the same focal plane at depth $z_f$, using the AOS light-field renderer~\cite{kurmi_aos_2018}. Everything the camera saw from all viewpoints is projected onto this plane and added up. Objects at the selected depth are seen from every viewpoint at the same place, so they add up. The occluder in front of the plane lands somewhere else in every viewpoint, so it gets averaged out.

\paragraph{Frame integration.}
For the frames we use the image-based rendering of AOS~\cite{kurmi_aos_2018}. Every pixel of every recorded image is one ray through the synthetic aperture. A virtual camera is defined with its own pose, field of view and focal plane. For each pixel of this virtual camera, a ray is cast through its projection center and intersected with the focal plane. All recorded rays that pass through this point on the focal plane are integrated and normalized, and the result is the color of that pixel. Points on the focal plane are hit by rays from every viewpoint at the same place, so they add up. An occluder in front of the plane blocks only some of the rays and lands somewhere else in every viewpoint, so it is averaged out if it is sparse enough and the aperture is wide and densely sampled~\cite{kurmi_aos_2018}. No 3D reconstruction of the scene is needed for this.

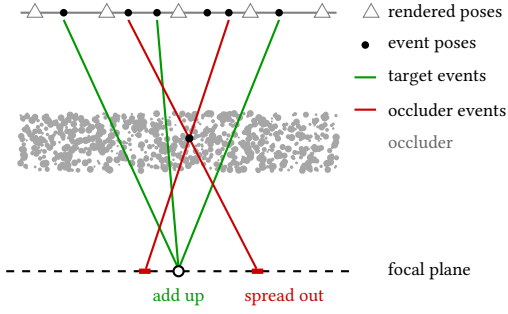
\begin{figure}[t]
\centering
\begin{tikzpicture}[font=\footnotesize, >={Stealth[length=1.6mm]}, scale=0.95,
  camlight/.style={regular polygon, regular polygon sides=3, draw=black!50, fill=white, inner sep=0pt, minimum size=2.4mm},
  ok/.style={green!60!black, thick},
  bad/.style={red!80!black, thick}]
  \begin{scope}
    \pgfmathsetseed{7}
    \foreach \i in {1,...,800} {
      \pgfmathsetmacro{\px}{rnd*4.4}
      \pgfmathsetmacro{\py}{1.4+rnd*0.8}
      \pgfmathsetmacro{\pr}{0.4+rnd*1.1}
      \fill[black!35] (\px,\py) circle (\pr pt);
    }
  \end{scope}
  \draw[dashed, thick] (-0.2,0) -- (4.6,0);
  \draw[black!50, thick] (0.0,3.6) -- (4.4,3.6);
  \foreach \x in {0.2,1.2,2.2,3.2,4.2} { \node[camlight] at (\x,3.6) {}; }
  \foreach \x in {0.6,1.5,1.9,2.6,2.9,3.6} { \fill (\x,3.6) circle (1.5pt); }
  \draw[ok] (0.6,3.5) -- (2.2,0.03);
  \draw[ok] (1.9,3.5) -- (2.2,0.03);
  \draw[ok] (3.6,3.5) -- (2.2,0.03);
  \draw[fill=white, thick] (2.2,0) circle (2pt);
  \coordinate (occ) at (2.35,1.85);
  \draw[bad] (1.5,3.5) -- (occ) -- ($(1.5,3.5)!2.1!(occ)$);
  \draw[bad] (2.9,3.5) -- (occ) -- ($(2.9,3.5)!2.1!(occ)$);
  \fill[black] (occ) circle (1.6pt);
  \draw[red!80!black, line width=2pt] (1.65,0) -- (1.81,0);
  \draw[red!80!black, line width=2pt] (3.22,0) -- (3.38,0);
  \node[green!60!black, below] at (2.2,-0.12) {add up};
  \node[red!80!black, below right] at (3.0,-0.12) {spread out};
  \begin{scope}[every node/.style={right, anchor=west}]
    \node at (5.0,3.6) {rendered poses};
    \node[camlight] at (4.8,3.6) {};
    \node at (5.0,3.15) {event poses};
    \fill (4.8,3.15) circle (1.5pt);
    \draw[ok] (4.65,2.7) -- (4.95,2.7);
    \node at (5.0,2.7) {target events};
    \draw[bad] (4.65,2.25) -- (4.95,2.25);
    \node at (5.0,2.25) {occluder events};
    \node[gray!80!black] at (5.0,1.8) {occluder};
    \node at (5.0,0) {focal plane};
  \end{scope}
\end{tikzpicture}
\caption{Event integration on the focal plane. Every event gets its own
pose between the rendered poses and is projected as one ray. Events from
the target on the focal plane land at the same point from every pose and
add up. Events from the occluder pass through the same point in front of
the plane, so they land at different places and spread out.}
\Description{Side-view schematic: a horizontal camera path on top with
rendered poses as triangles and event poses as dots between them, a band
of gray dots as the occluder, and a dashed focal plane at the bottom.
Three green rays from event poses meet at one point on the focal plane.
Two red rays from event poses cross at one point inside the occluder and
hit the focal plane at two separate places. A legend on the right names
the symbols.}
\label{fig:event-splat}
\end{figure}

\paragraph{Event integration.}
Events are sparse, asynchronous and have a sign, so there is no image to project. We therefore integrate every event on its own, in two steps (Figure~\ref{fig:event-splat}).

\emph{From events to rays.} Each event is stored with its pixel coordinates, its timestamp and polarity. The poses of the rendered frames are shared. An event fires while the camera moves, so its viewpoint is not one of the rendered poses. We estimate the pose of each event by interpolating between the two nearest poses. The position is interpolated linearly with the timestamp of the event, and the rotation along the shortest arc between the two rotations. This viewpoint is only as fine as the timestamps v2e assigns within a frame interval (Section~\ref{sec:eventmodel}), but it is still much finer than the rendered poses. Using the nearest rendered pose instead would collapse all events of one interval onto one viewpoint, and at high speed the aperture would then be sampled in coarse steps. The direction of the ray is given by the position of the pixel on the sensor. The offset of the pixel from the image center, divided by the focal length of the camera, gives the angle of the ray. A shorter focal length therefore spreads the same pixels over a wider angle, and each pixel looks in a slightly different direction. The focal length and the image center are the same for all events and are taken from the rendered camera. For an event at pixel $(x, y)$ the direction in the camera frame is
\[
  \mathbf{d} = \left(\frac{c_x - x}{f_x},\ \frac{c_y - y}{f_y},\ -1\right),
\]
with the principal point $(c_x, c_y)$ and the focal length $(f_x, f_y)$ in pixels. The ray is then rotated into the world with the interpolated pose. This gives a ray in world space, together with the directions that span the footprint of one sensor pixel.

\emph{Splatting on the focal plane.} Each ray is intersected with the focal plane and the footprint of its pixel is projected there. Seen from the side, this footprint is a distorted quadrilateral. The event deposits its polarity over this footprint with a tent weighting that is widened to twice the footprint. Neighboring events overlap on the plane and their weights sum to one, so the sparse events form a smooth surface. The weight of each event is divided by the projected area of the footprint. An event far from the optical axis, whose footprint projects larger, therefore has the same total weight as a central one. All splats are added up on the GPU, with ON and OFF polarity in separate channels. One quad is drawn per event, so the cost grows with the number of events. The rays do not depend on the focal depth, so refocusing to another $z_f$ reuses all rays and only needs a redraw. After that, the final reconstruction counts the positive and negative events in each pixel, and we have two ways of visualizing the result. The first one takes the positive minus the negative count and maps it to gray, so that pixels where both polarities cancel are mid-gray. The second one takes the absolute value of this difference and inverts it, so cancelled pixels become white and pixels where one polarity dominates become dark.

\begin{figure}[t]
\centering
\begin{minipage}[t]{0.48\columnwidth}
\centering
\includegraphics[width=\linewidth]{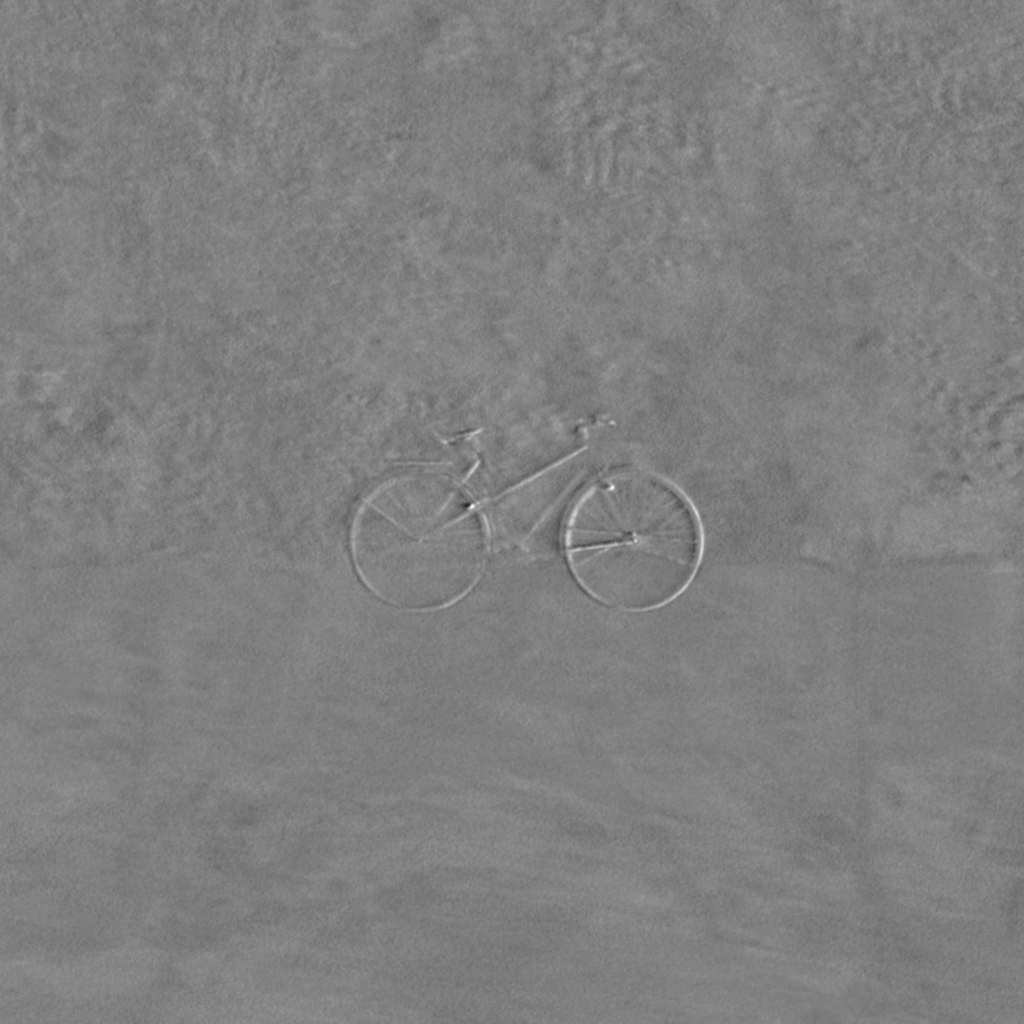}\\[0.3em]
{\small (a) Signed net polarity}
\end{minipage}\hfill
\begin{minipage}[t]{0.48\columnwidth}
\centering
\includegraphics[width=\linewidth]{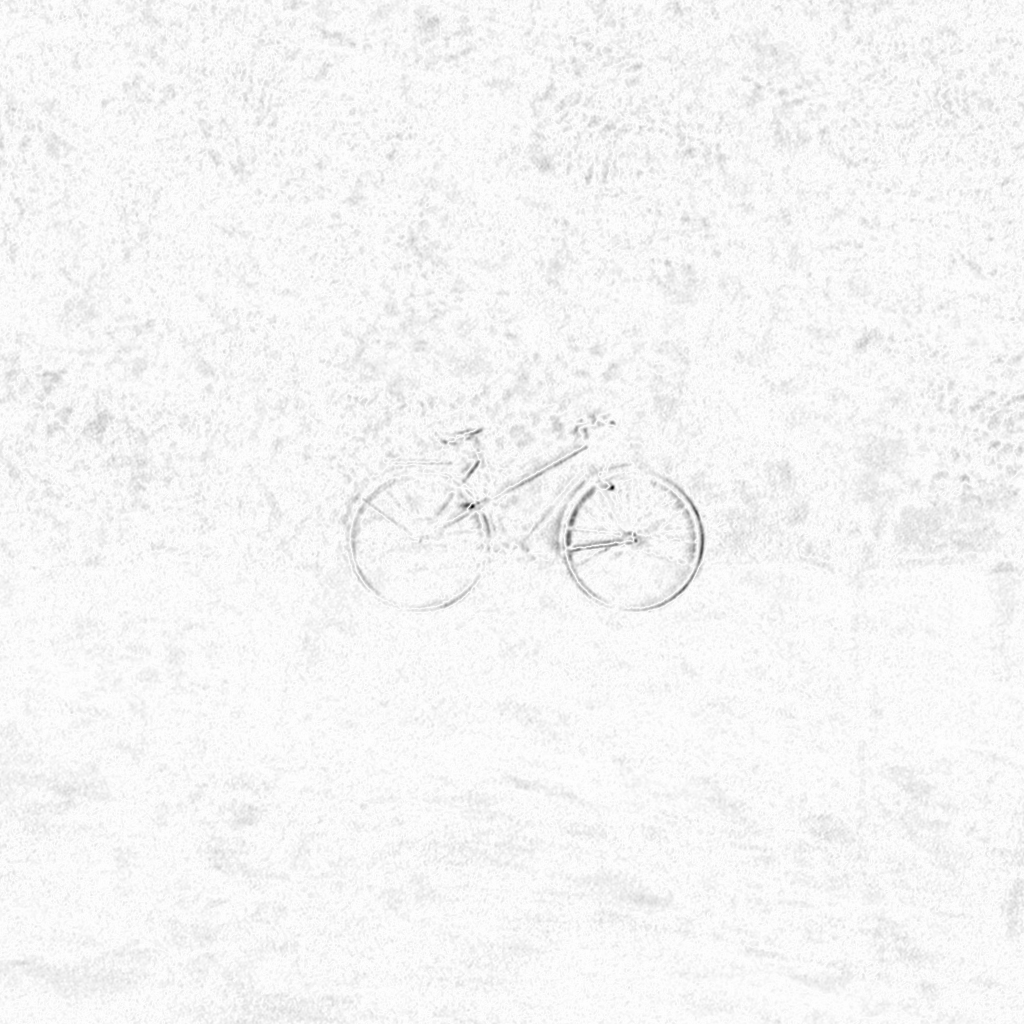}\\[0.3em]
{\small (b) Absolute net polarity}
\end{minipage}
\caption{Event-based synthetic-aperture integral produced by the proposed
integration path (bright scene, $25$\,m/s). (a)~Signed net polarity,
balanced pixels render mid-gray. (b)~Absolute value of the net polarity,
inverted, so cancelled pixels render white and events black. The occluding foliage of
Figure~\ref{fig:scene} washes out, while the bicycle on the focal
plane, including spokes and frame, is reinforced.}
\Description{Two square grayscale reconstructions side by side: on the
left a clearly outlined bicycle with visible spokes on a nearly uniform
gray background, on the right the same bicycle in dark gray on a white
background; the foliage visible in the earlier scene figure is gone in
both.}
\label{fig:event-sai-demo}
\end{figure}

Figure~\ref{fig:event-sai-demo} shows a reconstruction produced by this
path: integrated over the full trajectory, the signed splats from the
near-field foliage cancel toward mid-gray while the bicycle on the focal
plane accumulates coherently.

\subsection{Event-Rate Controller}
\label{sec:erc}
A frame camera with a given frame rate always puts out the same amount of data. For event cameras it depends on the light and also very much on the scene itself~\cite{gallego_survey_2022}. For scenes where the camera is static, only very few events are produced. However, for scenes where the camera is moving and a lot of texture is captured, a lot of events are recorded. In such scenarios it can happen that the event camera produces more events than the link from the event camera can handle, especially for sensors with higher resolution like the one we are simulating in our experiments~\cite{gehrig_high_res_2022, finateu_imx636_2020}. In older systems, when the link is saturated, events are dropped in an uncontrolled manner, which leads to signal loss, and the timestamps of the events get corrupted~\cite{gallego_survey_2022, gehrig_high_res_2022}. Modern systems ship with an event-rate controller (ERC) that drops events in a controlled manner and can be configured by setting an event rate per second~\cite{finateu_imx636_2020}. In our high-speed scene the simulated event stream exceeds the link budget we estimate in Section~\ref{sec:erc-budget}, so a real recording of this scenario would likely have the ERC active, and simulating the camera without ERC would be too optimistic.

However, how exactly the ERC decides which events to drop is not published, so we follow the description of Gehrig and Scaramuzza~\cite{gehrig_high_res_2022} and use a uniform dropping algorithm. The event stream is split into short counting periods, and whenever a period contains more events than the configured rate allows, the surplus is dropped at random. The event rate and the counting period we use are given in Section~\ref{sec:setup}.

\section{Experimental Setup}
\label{sec:setup}

\begin{figure}[t]
\centering
\includegraphics[width=0.92\columnwidth]{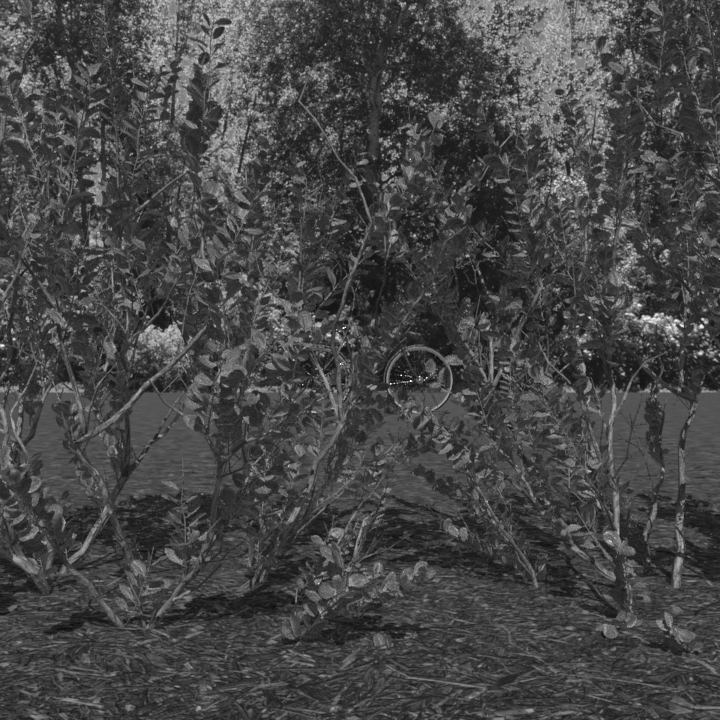}
\caption{One viewpoint from the shared trajectory. Near-field foliage
occludes a bicycle on the focal plane; the $0.5$\,m path supplies
the remaining viewpoints of the synthetic aperture.}
\Description{Grayscale rendered view through two bushes. A bicycle wheel is
partly visible in a gap in the foliage; the ground is covered in leaves.}
\label{fig:scene}
\end{figure}

The comparison uses a scene with two bushes as occluders and a bicycle behind them (Figure~\ref{fig:scene}). The bicycle is about $9$\,m away from the camera. The camera moves $0.5$\,m along the occluders and rises slightly while doing so, which makes the path a diagonal shift without any rotation. Although our event integration works with arbitrary 6-DoF poses, we keep to this simple movement, because Bimber et al.~\cite{bimber_robot_dogs_2025} found that a lateral shift suppresses occluders better than a rotation, as it maximizes the parallax, and that a vertical component on top of it also helps against horizontally aligned occluders. This movement is what is rendered. We chose this rather short aperture on purpose. With a long aperture of several meters both sensors get more than enough views, the two results converge, and the differences we want to show disappear. The short aperture is where a shortage of views actually hurts. Our simulation is also optimistic, because the poses are known exactly. A real system has pose errors and needs even more views to average them out, so the situation we create here with a short aperture would show up on a real system at larger apertures too. We render 3000 frames for this movement to make high \texttt{cutoff\_hz} settings possible. This render is only done once. The scene has one global light in Blender. The resolution of the Blender render is $720\times720$ pixels. We compare the sensors at the same resolution to make bandwidth comparisons fair.

\subsection{Two Operating Points}
\label{sec:studies}
The comparison is done at two operating points and one additional run, listed in Table~\ref{tab:studies}. The first operating point is a bright high-speed scenario, where the speed of the camera is $25$\,m/s. Here we want to see if the event camera can still produce a good reconstruction when the frame camera is starved of frames, because at $160$\,fps the frame camera only records three frames during the movement. The second operating point is a low-light scenario. The movement here is much slower at $5$\,m/s, and therefore the frame camera has enough frames to create a good result in principle, so this is all about sensor noise in a low-lit environment. We use $5$\,lux, because this is the lowest illuminance for which Prophesee, the manufacturer of the event sensor, still specifies a latency value~\cite{prophesee_gen41}. The scene, the path and the render are the same at both comparisons. A third run in the bright scene at $1$\,m/s is used to compare reconstruction quality at two approximately matched data budgets (Section~\ref{sec:res-bandwidth}). We pair an ERC cap of $20$\,MEv/s with $80$ frames at $160$\,fps, and $10$\,MEv/s with $40$ frames at $80$\,fps. The $40$-frame sequence takes every second frame of the original $80$-frame sequence, preserving the exposure and noise realizations. Both pairs cover the same $0.5$\,s traversal; only the frame sampling rate and the ERC cap change.

\begin{table}[t]
\caption{The two evaluation points and the additional run at $1$\,m/s. Frame-view counts are
$n = 160\,\text{fps}\cdot T$ over the $0.5$\,m aperture, rounded down.}
\label{tab:studies}
\begin{tabular}{@{}lccc@{}}
\toprule
 & High speed & Low light & Matched data \\
\midrule
Scene illuminance & $1000$\,lux & $5$\,lux & $1000$\,lux \\
Camera speed & $25$\,m/s & $5$\,m/s & $1$\,m/s \\
Duration $T$ & $0.02$\,s & $0.1$\,s & $0.5$\,s \\
Frame views $n$ & $3$ & $16$ & $80$ / $40$ \\
Event \texttt{cutoff\_hz} & $1000$\,Hz & $100$\,Hz & $1000$\,Hz \\
ERC (MEv/s) & $50$ / $20$ & off & $20$ / $10$ \\
\bottomrule
\end{tabular}
\end{table}

\subsection{Sensor Parameters}
\label{sec:params}
We model two commercially available industrial sensors. The frame camera is a Basler ace~2 with the Sony IMX392, a global-shutter CMOS sensor. The event camera is the Sony IMX636, a current Prophesee sensor. Their parameters are collected in Tables~\ref{tab:frame-params} and~\ref{tab:event-params}. Most of them are manufacturer data. The rest of this section explains the values we had to choose ourselves.

\begin{table}[t]
\caption{Frame-camera model. EMVA rows are Basler's measurement of the
ace~2 a2A1920-160umBAS (Sony IMX392)~\cite{basler_emva_overview}.
Exposures are chosen as the frame camera's best case at each operating
point.}
\label{tab:frame-params}
\small
\begin{tabular}{@{}lrl@{}}
\toprule
Parameter & Value & Source \\
\midrule
Simulated resolution & $720\times720$ & matched \\
Frame rate & $160$\,fps & EMVA \\
Peak QE & $62\%$ & EMVA \\
Pixel pitch & $3.45$\,\textmu m & EMVA \\
Read noise & $3.0$\,e$^-$ & EMVA \\
Full-well capacity & $10.5$\,ke$^-$ & EMVA \\
Exposure (bright / night) & $0.1$ / $6.25$\,ms & chosen \\
\bottomrule
\end{tabular}
\end{table}

\begin{table}[t]
\caption{Event-camera model. IMX636-class target
specifications~\cite{finateu_imx636_2020, prophesee_gen41} and v2e
non-ideality settings. The photoreceptor cutoff is a modelling
assumption (Section~\ref{sec:cutoff}), not a datasheet number.}
\label{tab:event-params}
\small
\begin{tabular}{@{}lrl@{}}
\toprule
Parameter & Value & Source \\
\midrule
Simulated resolution & $720\times720$ & matched \\
ERC budget (bright) & $50$\,MEv/s & Section~\ref{sec:erc-budget} \\
Contrast threshold & $0.25$ & brief \\
Threshold mismatch & $0.03$ & v2e \\
Refractory period & $0.1$\,ms & v2e \\
Shot-noise rate (bright / night) & $0.1$ / $10$\,Hz/px & brief \\
Leak rate & $0.1$\,Hz/px & brief \\
\texttt{cutoff\_hz} (bright / night) & $1000$ / $100$\,Hz & Section~\ref{sec:cutoff} \\
\bottomrule
\end{tabular}
\end{table}

\paragraph{Frame camera.}
Quantum efficiency, pixel pitch, read noise, full-well capacity and frame rate come from Basler's published EMVA~1288 measurement~\cite{basler_emva_overview}. The exposure can never be longer than the $6.25$\,ms frame interval. At night we use this full interval, so the frame camera collects as much light as a $160$\,fps global-shutter sensor can. In the bright scenario we use a $0.1$\,ms exposure. The image moves about $1.5$\,px/ms at $25$\,m/s, so the motion blur stays around $0.15$\,px, and the result is limited by the three frames on the path and not by blur or noise. These choices are the best case for the frame camera at each operating point.

\paragraph{Event camera.}
The contrast threshold of $0.25$ is the nominal $25\%$ logarithmic contrast from the product brief~\cite{prophesee_gen41}. Threshold mismatch and refractory period are the settings v2e recommends~\cite{hu_v2e_2021}. For the shot noise the brief gives $0.1$\,Hz per pixel at $1$\,klux and allows up to $10$\,Hz per pixel at $5$\,lux, and we use these two values for the bright and the night scenario. The photoreceptor cutoff has no datasheet value and is treated in Section~\ref{sec:cutoff}.

\subsection{Photoreceptor Cutoff}
\label{sec:cutoff}
To our knowledge there is no published measurement of the photoreceptor bandwidth of the IMX636 at a known illuminance~\cite{mcreynolds_step_2024}, so we cannot look up \texttt{cutoff\_hz} and have to choose it. We base the choice on two sources.

The first one is the original DVS. Its bandwidth was measured directly: about $3$\,kHz at $1$\,klux and about $300$\,Hz at a thousand times less light~\cite{lichtsteiner_dvs_2008, gallego_survey_2022}. That sensor had much larger pixels than the IMX636. Smaller pixels collect less light and get slower in the dark~\cite{gehrig_high_res_2022}, so the IMX636 should have lower cutoff values at the same light levels.

The second one is the product brief of the IMX636. It gives the latency of the sensor: $100$\,\textmu s at $1$\,klux and $1000$\,\textmu s at $5$\,lux~\cite{prophesee_gen41}. We use the latency as the time constant of the cutoff filter, so the cutoff is $f_c = 1/(2\pi\,t_{\text{lat}})$. This is an assumption, since the latency also depends on the contrast of the step. With a higher contrast the pixel fires a little earlier, but the datasheet does not say which contrast was used. With it, the two latencies give $1.6$\,kHz and $160$\,Hz.

We stay below both sources and use $1000$\,Hz for the bright scenario and $100$\,Hz for the night scenario. The lux values of the brief are the manufacturer's test conditions, and we take them as our scene illuminances.

\subsection{Event-Rate Budget}
\label{sec:erc-budget}
The ERC model is described in Section~\ref{sec:erc}. Here we give the two values it needs, the event rate and the counting period.

The event rate follows from how fast the camera can get its events to the computer. The IMX636 is sold in Prophesee's EVK4 camera, which is connected over USB~3.0 and has a maximum bandwidth of $1.6$\,Gbit/s~\cite{prophesee_evk4}. How many events fit through this link depends on the size of one event. The sensor packs its events into the EVT3 format, which only transmits what changes between neighboring events, so the size of an event depends on the scene~\cite{prophesee_evt3}. We estimate the data rate using $4$ bytes per event. This is a working approximation rather than a strict upper bound for EVT3. For context, Prophesee's published RAW EVT3 recordings yield approximately $2.8$ bytes per event for high-speed laser motion, $3.2$ for a moving-camera driving sequence and $4.7$ for pedestrians, calculated from their listed file sizes, durations and event rates~\cite{prophesee_sample_recordings}. Using our four-byte estimate, the link supports
\[
  1.6 \times 10^{9} / (8 \times 4) = 50\,\text{MEv/s}.
\]
To be safe we also run the ERC at $20$\,MEv/s, well below any plausible link rate. Events are counted in periods of $200$\,\textmu s, the reference period of the IMX636 ERC~\cite{finateu_imx636_2020}. The bright stream at $25$\,m/s is about $177$\,MEv/s and therefore exceeds the estimated link budget by about $3.5\times$, which motivates the ERC experiments. That reconstruction is repeated after the ERC at both rates. The night $5$\,m/s stream at the $100$\,Hz cutoff is about $21$\,MEv/s, under budget, and is left as it is.

\section{Results}
\label{sec:results}

Reconstruction quality is assessed qualitatively. We look at whether the bicycle can be recognized, which structures are visible, such as the spokes, the frame and the foliage that should be washed out, and how noisy the image is. The conclusions apply to the two operating points in Table~\ref{tab:studies} and to the additional run at $1$\,m/s, and not to the speeds and illuminances between them. Section~\ref{sec:res-bandwidth} then compares the data the two cameras produce in all three runs.

\subsection{High Speed}
\label{sec:res-fast}
\begin{figure}[t]
\centering
\begin{minipage}[t]{0.48\columnwidth}
\centering
\includegraphics[width=\linewidth]{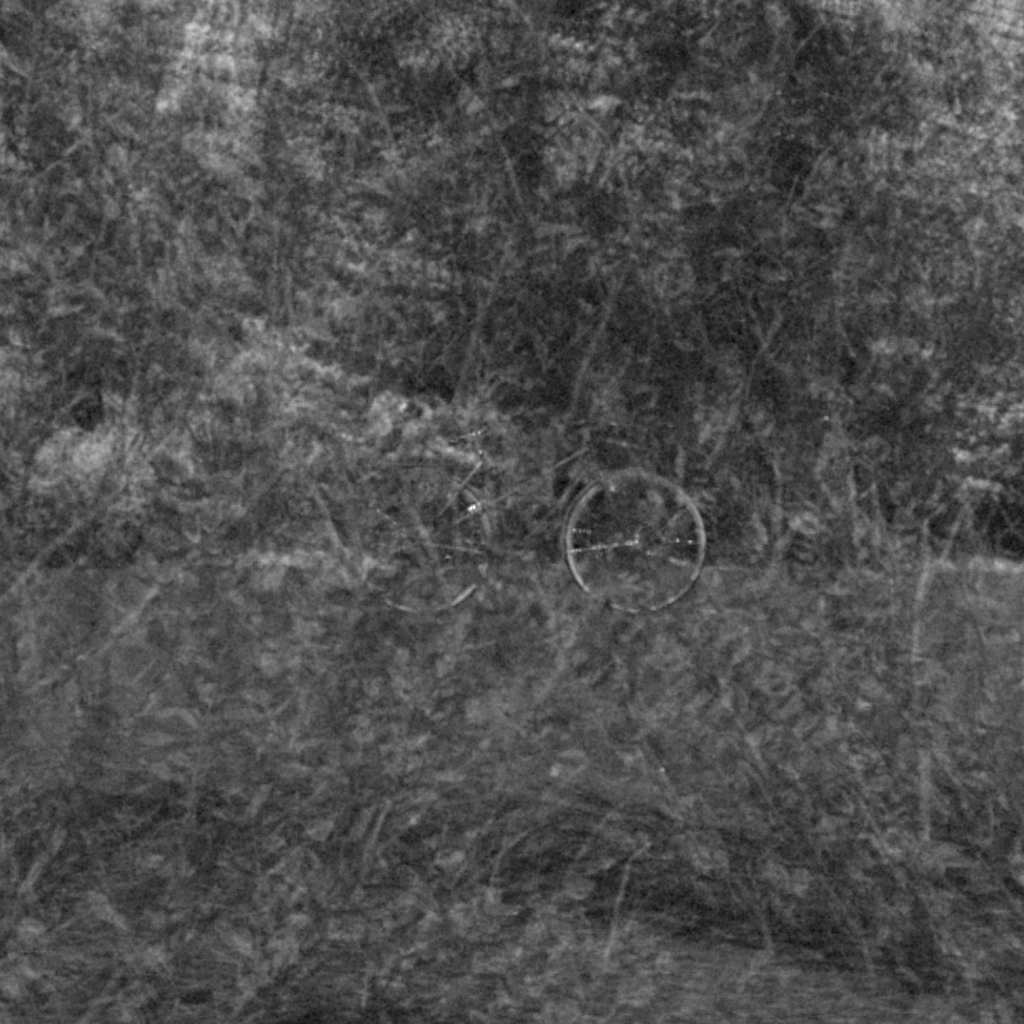}\\[0.3em]
{\small (a) Frame SAI, $3$ views}
\end{minipage}\hfill
\begin{minipage}[t]{0.48\columnwidth}
\centering
\includegraphics[width=\linewidth]{assets/res_fast_event.png}\\[0.3em]
{\small (b) Event SAI, full stream}
\end{minipage}\\[0.6em]
\begin{minipage}[t]{0.48\columnwidth}
\centering
\includegraphics[width=\linewidth]{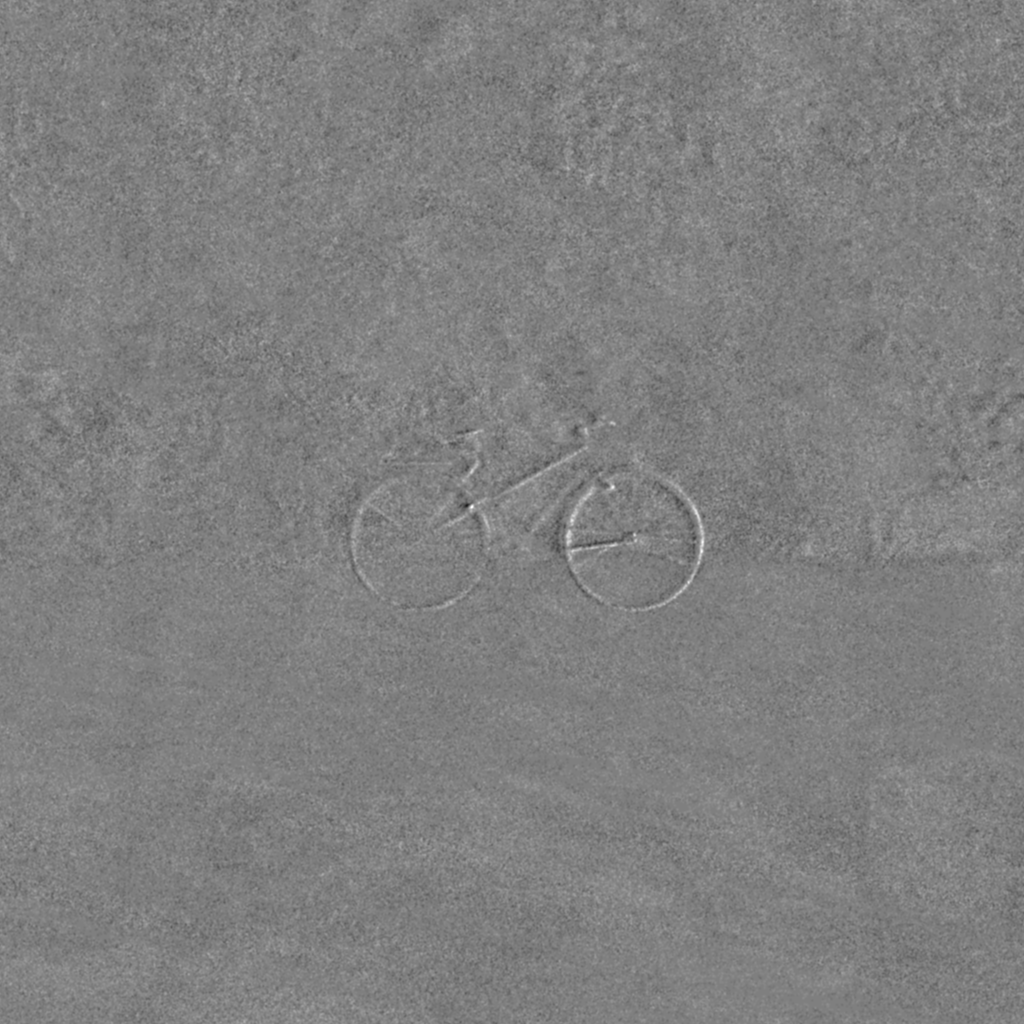}\\[0.3em]
{\small (c) Event SAI, ERC $50$\,MEv/s}
\end{minipage}\hfill
\begin{minipage}[t]{0.48\columnwidth}
\centering
\includegraphics[width=\linewidth]{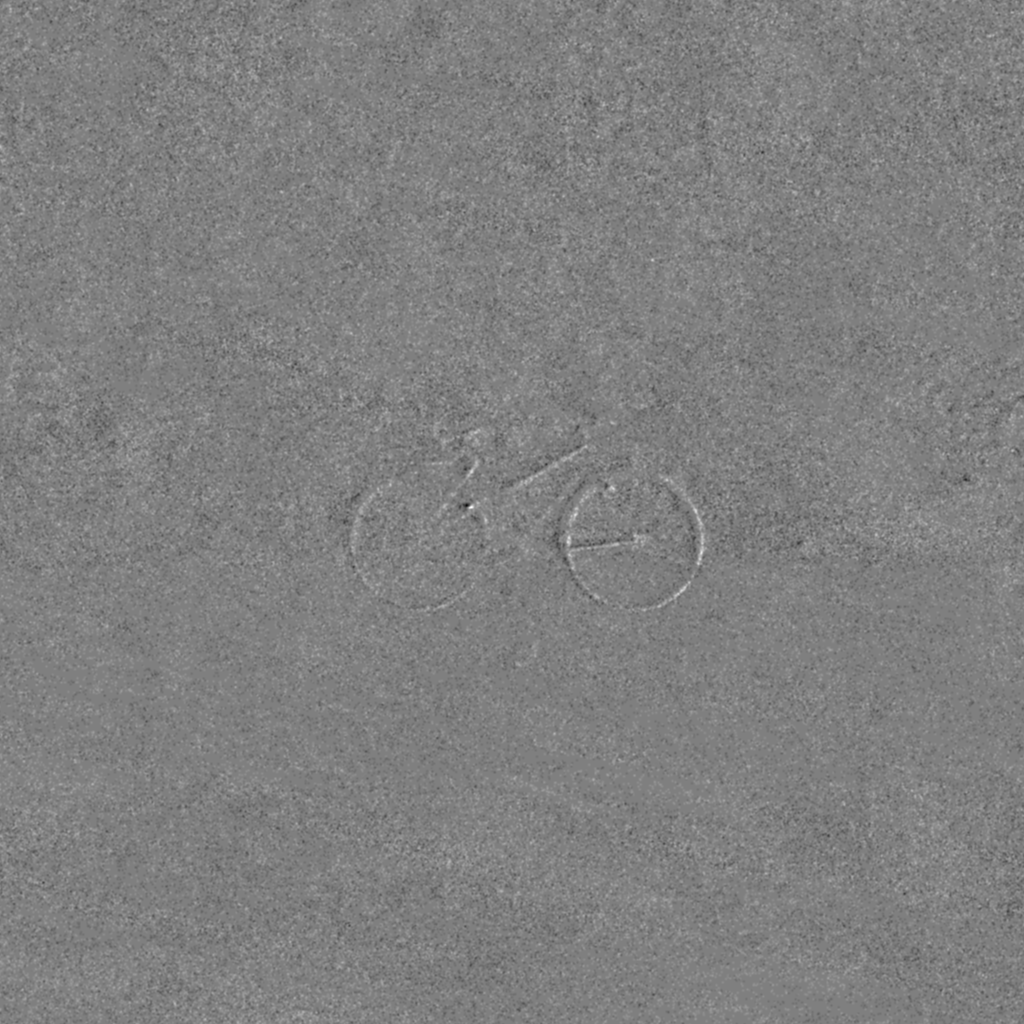}\\[0.3em]
{\small (d) Event SAI, ERC $20$\,MEv/s}
\end{minipage}
\caption{High speed ($25$\,m/s, $1000$\,lux). (a)~Frame SAI from the three frames the camera records on the path. (b)~Event SAI from the full event stream. (c)~and (d)~Event SAI after the ERC drops the stream to $50$ and $20$\,MEv/s. The bicycle survives both; fine structure and smoothness do not.}
\Description{Four square grayscale reconstructions. Top left: a wall of foliage in which only the two wheel rims of the bicycle can be made out. Top right: a clearly outlined bicycle with spokes on a nearly uniform gray background. Bottom: the same bicycle after capping, both grainier with fainter spokes, the right one grainier than the left.}
\label{fig:res-fast}
\end{figure}

At $25$\,m/s and $1000$\,lux the frame camera records three frames on the path. Figure~\ref{fig:res-fast} shows both reconstructions. The three images the frame camera produces are not enough to create a good reconstruction. Some parts of the bike are still visible, but that is only because the occlusion is not very dense, as shown in Figure~\ref{fig:scene}. The input frames are well exposed and do not have any noise problems. The event camera at the same time captures $3.5$\,M events along the same path, and when looking at its result it is clear that the event reconstruction gives us the whole bike. However, the raw event data we get at this combination of speed and \texttt{cutoff\_hz} is above the link budget we estimated in Section~\ref{sec:erc-budget}. Therefore we also added two reconstructions after an event-rate controller simulation. When the ERC limit is set to $50$\,MEv/s, about two thirds of the events are dropped, but the reconstruction still shows the bike, at the cost of detail and noise. We also did the same with an event budget of $20$\,MEv/s. The bike is still recognizable, but the quality degrades further and there is less detail. The data behind these images is compared in Section~\ref{sec:res-bandwidth}.

\subsection{Low Light}
\label{sec:res-night}
\begin{figure}[t]
\centering
\begin{minipage}[t]{0.48\columnwidth}
\centering
\includegraphics[width=\linewidth]{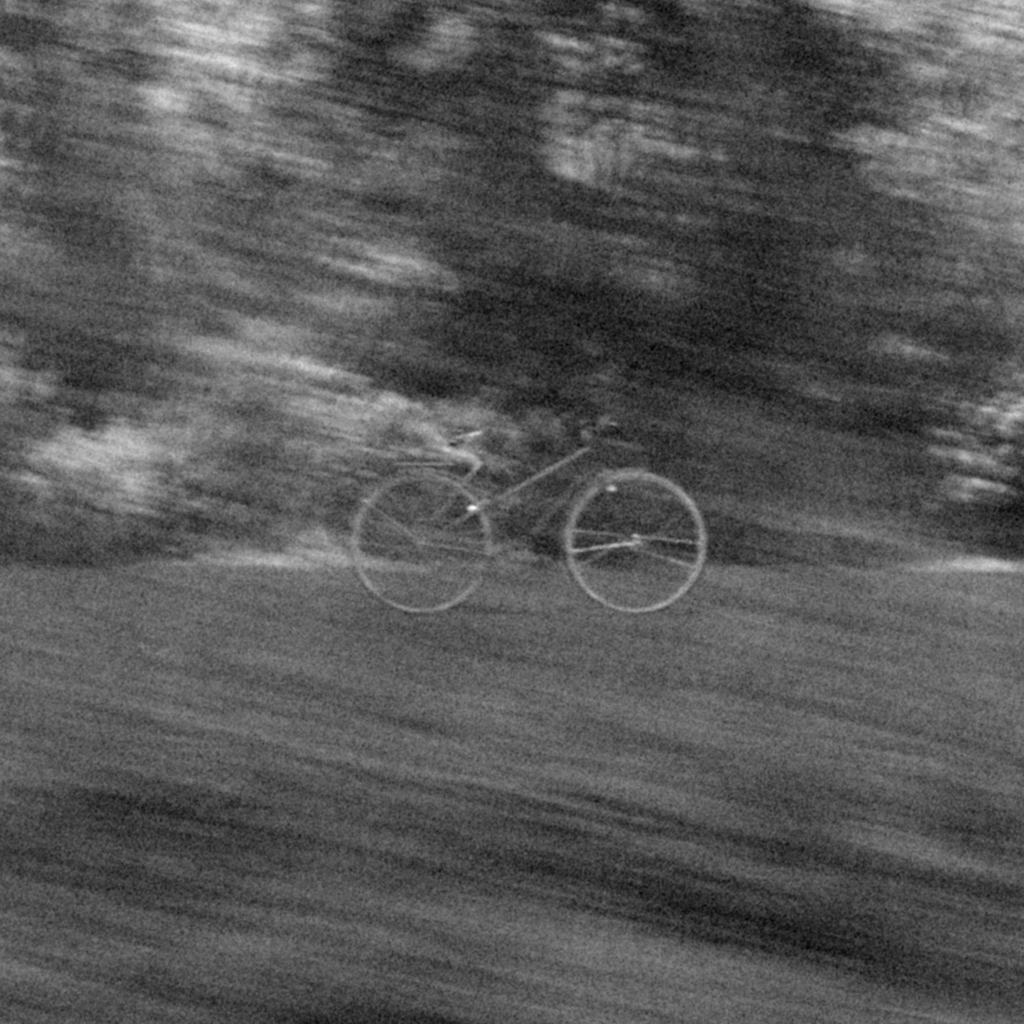}\\[0.3em]
{\small (a) Frame SAI, $16$ views}
\end{minipage}\hfill
\begin{minipage}[t]{0.48\columnwidth}
\centering
\includegraphics[width=\linewidth]{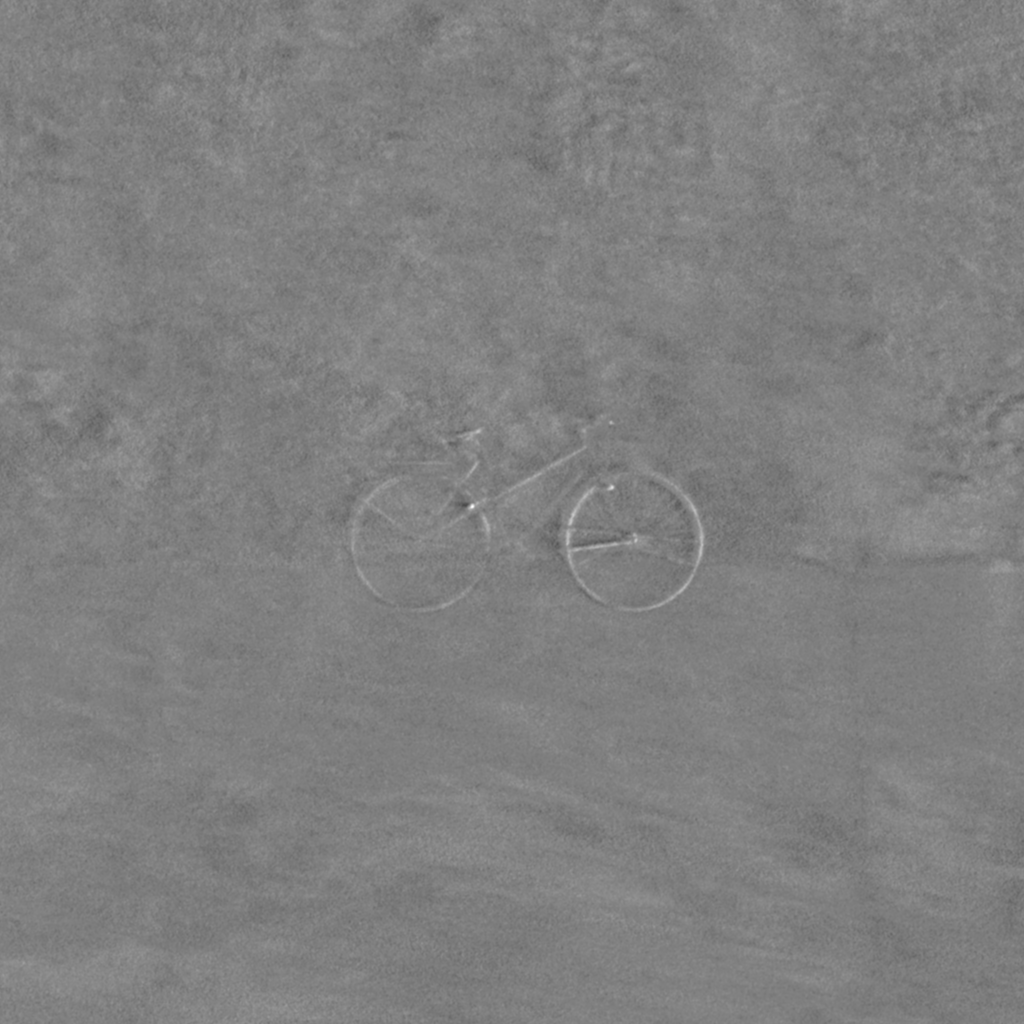}\\[0.3em]
{\small (b) Event SAI, $100$\,Hz cutoff}
\end{minipage}
\caption{Low light ($5$\,m/s, $5$\,lux). (a)~Frame SAI from $16$ views. (b)~Event SAI at the $100$\,Hz photoreceptor cutoff. Both recover the bicycle at a similar quality.}
\Description{Two square reconstructions. The left (frame) image shows a
noisy bicycle with streaked leftover foliage. The right (event) image
shows a fainter but complete bicycle on a smoother gray field.}
\label{fig:res-night}
\end{figure}

At $5$\,lux and $5$\,m/s the frame camera records $16$ views and exposes for the full $6.25$\,ms between two frames. The event camera runs with the $100$\,Hz cutoff of Section~\ref{sec:cutoff}. Both reconstructions (Figure~\ref{fig:res-night}) recover the bicycle at a comparable quality. The frame side looks a little bit more noisy, but the level of detail of the bike is very similar. In our simulation we cannot claim an advantage for either.

\subsection{Bandwidth}
\label{sec:res-bandwidth}
We distinguish the data rate in MB/s, which determines whether a stream fits through the camera link, and the data volume that is used as input for one reconstruction. Table~\ref{tab:bandwidth} reports both, using the estimate of $4$ bytes per event from Section~\ref{sec:erc-budget} and $8$ bits per pixel for the frames. These are payload estimates.

The raw high-speed event stream produces about $709$\,MB/s, exceeding the estimated $200$\,MB/s link budget. At an ERC cap of $20$\,MEv/s, one traversal produces approximately $1.6$\,MB, comparable to the three frames, while the bicycle remains more recognizable in the event reconstruction (Figure~\ref{fig:res-fast}). At night the two streams produce similar volumes, $8.6$\,MB for events and $8.3$\,MB for frames. The raw bright stream at $1$\,m/s fits within the estimated link budget, but its $86.7$\,MB per traversal exceeds the $41.5$\,MB of the $80$ frames.

\begin{figure}[t]
\centering
\begin{minipage}[t]{0.48\columnwidth}
\centering
\includegraphics[width=\linewidth]{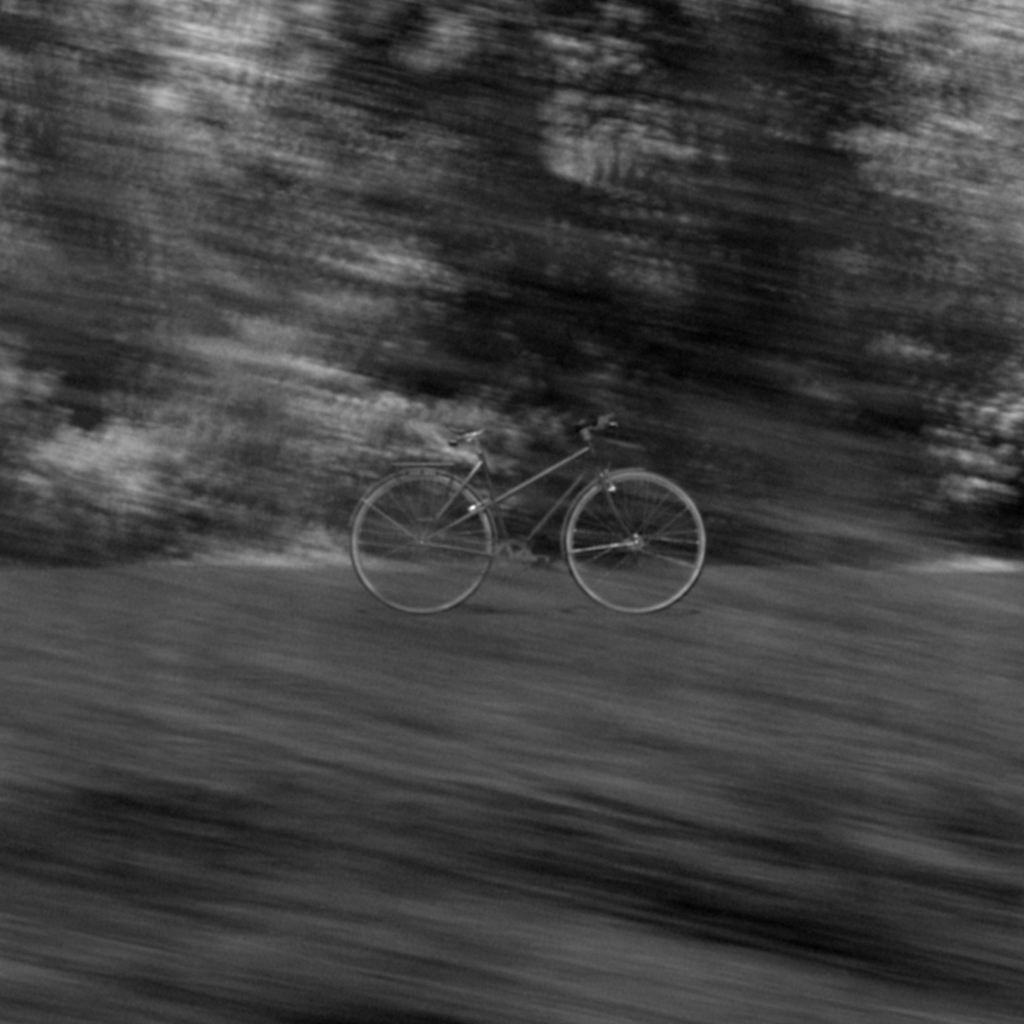}\\[0.3em]
{\small (a) Frames, $160$\,fps\\$80$ views, $41.5$\,MB}
\end{minipage}\hfill
\begin{minipage}[t]{0.48\columnwidth}
\centering
\includegraphics[width=\linewidth]{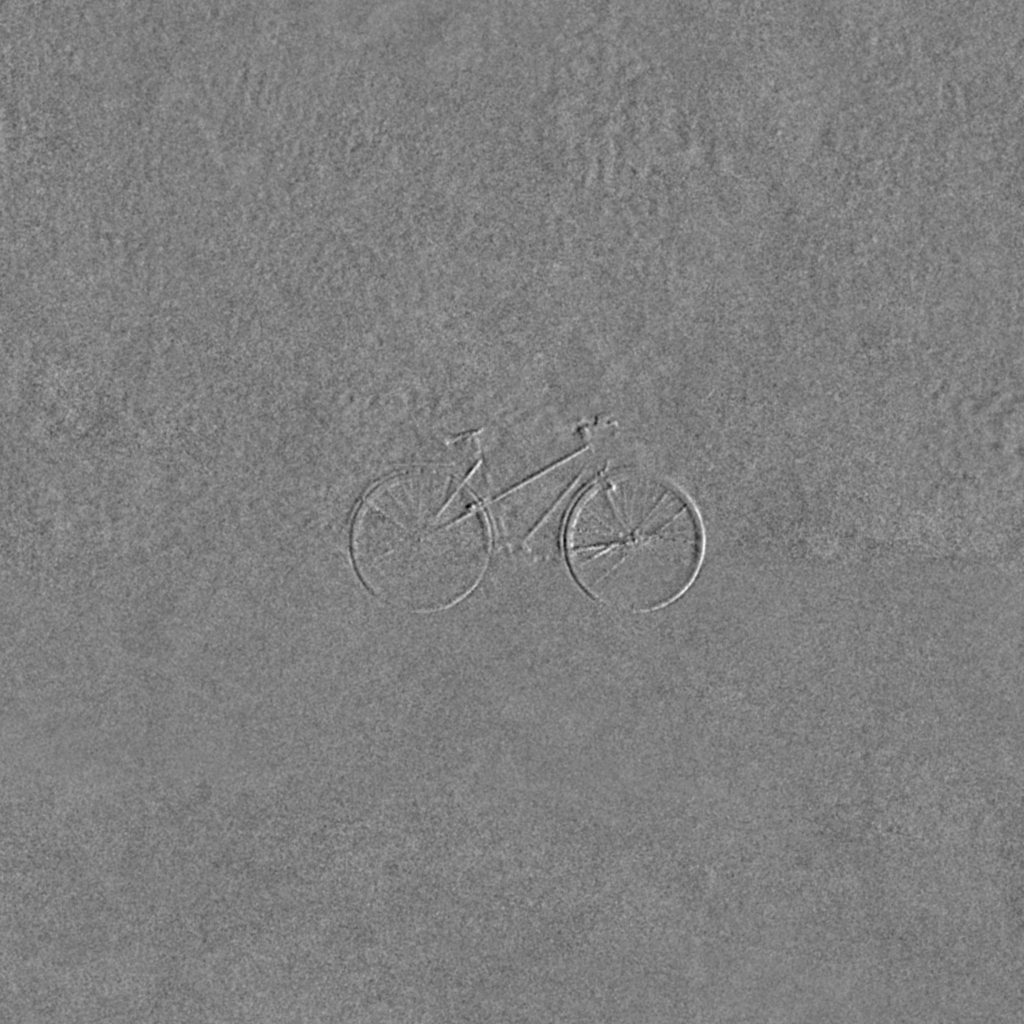}\\[0.3em]
{\small (b) Events, ERC $20$\,MEv/s\\$39.6$\,MB}
\end{minipage}\\[0.6em]
\begin{minipage}[t]{0.48\columnwidth}
\centering
\includegraphics[width=\linewidth]{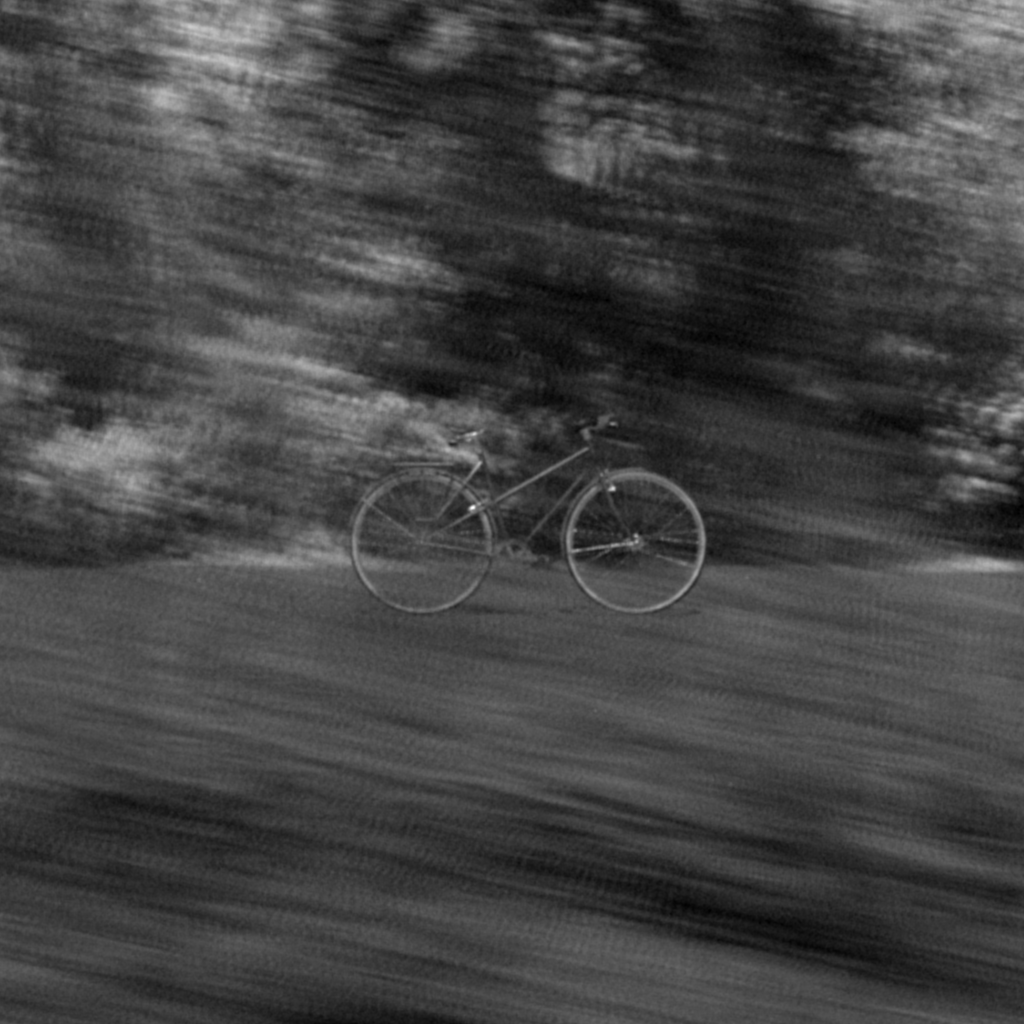}\\[0.3em]
{\small (c) Frames, $80$\,fps\\$40$ views, $20.7$\,MB}
\end{minipage}\hfill
\begin{minipage}[t]{0.48\columnwidth}
\centering
\includegraphics[width=\linewidth]{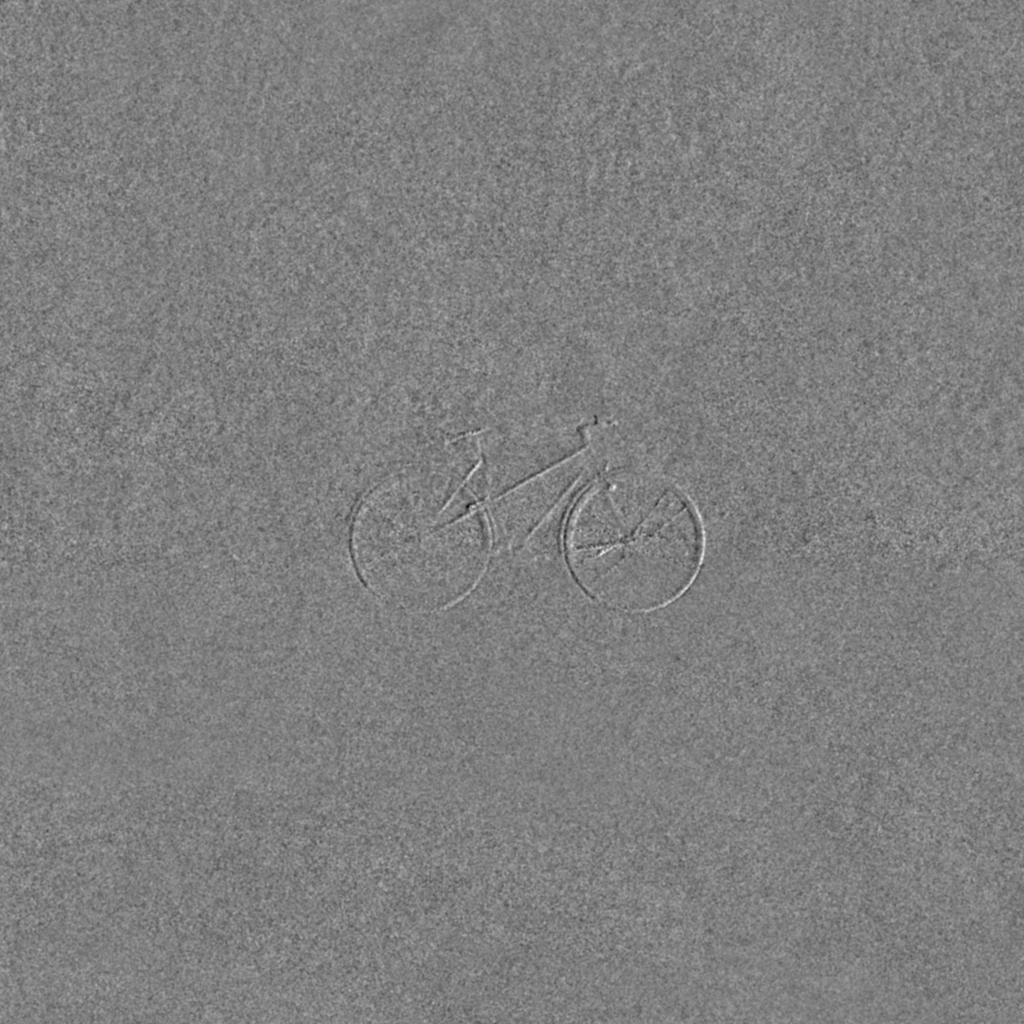}\\[0.3em]
{\small (d) Events, ERC $10$\,MEv/s\\$19.8$\,MB}
\end{minipage}
\caption{Two approximately matched data budgets in the bright scene at $1$\,m/s. Top: $80$ frames versus an ERC cap of $20$\,MEv/s. Bottom: $40$ frames versus $10$\,MEv/s. Volumes refer to one $0.5$\,s traversal. The $40$ views are every second frame of the original sequence, with unchanged exposure. Both modalities retain the bicycle at both budgets.}
\Description{Four grayscale reconstructions arranged by data budget in rows and sensor modality in columns. Left: the 80-frame and 40-frame reconstructions both show the bicycle against streaked residual foliage. Right: the 20 and 10 MEv/s event reconstructions show the bicycle against a more uniform but grainy background; the lower-budget image has more grain and less complete fine edges.}
\label{fig:res-data}
\end{figure}

For the bright $1$\,m/s run, we therefore reduce the data on both sides and compare two approximately matched budgets (Figure~\ref{fig:res-data}). The first pair uses the ERC at $20$\,MEv/s and $80$ frames at $160$\,fps, producing $39.6$ and $41.5$\,MB per traversal. The second pair halves both settings: the ERC at $10$\,MEv/s produces $19.8$\,MB, and $40$ frames at $80$\,fps produce $20.7$\,MB.

Event cameras are often seen as very efficient and less bandwidth hungry~\cite{gallego_survey_2022}, although some research already shows that this is not the case for all scenarios~\cite{gehrig_high_res_2022}. With a stationary camera, events come only from the moving targets, while with a moving camera the whole static scene triggers events~\cite{gallego_survey_2022}. SAI is exactly the second case, a moving camera and a stationary scene. In addition, SAI is all about occluders, and occluders have a lot of texture that produces a lot of events, because the event rate grows with the image gradient and the speed on the sensor~\cite{gallego_survey_2022}. In the $1$\,m/s comparison with matched budgets the event camera has no advantage over the frame camera in terms of reconstruction quality. Especially in the $10$\,MEv/s case the frame camera retains almost all of the detail with $40$ frames, while the event camera loses some detail compared to the higher budget.

The only bandwidth advantage we could see was at very high speed. At slower speeds this advantage seems to vanish in our simulation.

\begin{table}[t]
\caption{Data of the two cameras in all runs at $720\times720$. Event rates are measured from the simulated streams and converted with $4$ bytes per event (Section~\ref{sec:erc-budget}). Frames are $8$\,bit per pixel at $160$\,fps, except for the $40$-view subset at $80$\,fps. The last column is the data of one traversal of the $0.5$\,m aperture.}
\label{tab:bandwidth}
\small
\begin{tabular}{@{}llrrr@{}}
\toprule
Run & Stream & MEv/s & MB/s & MB/trav. \\
\midrule
$25$\,m/s, $0.02$\,s & Frames ($3$) & --- & $83$ & $1.6$ \\
 & Events, raw & $177$ & $709$ & $14.1$ \\
 & Events, ERC $50$ & $50$ & $200$ & $4.0$ \\
 & Events, ERC $20$ & $20$ & $80$ & $1.6$ \\
\midrule
$5$\,m/s, $0.1$\,s, night & Frames ($16$) & --- & $83$ & $8.3$ \\
 & Events, raw & $21$ & $85$ & $8.6$ \\
\midrule
$1$\,m/s, $0.5$\,s & Frames ($80$) & --- & $83$ & $41.5$ \\
 & Events, raw & $43$ & $174$ & $86.7$ \\
 & Events, ERC $20$ & $19.8$ & $79.3$ & $39.6$ \\
 & Frames ($40$) & --- & $41.5$ & $20.7$ \\
 & Events, ERC $10$ & $9.9$ & $39.7$ & $19.8$ \\
\bottomrule
\end{tabular}
\end{table}

\section{Discussion}
\label{sec:discussion}

\paragraph{What the three runs imply.}
The high-speed run shows the clearest difference. The deficit of the frame camera there cannot be resolved by adjusting the exposure. At the tested speed and aperture length, more viewpoints are needed, which the event camera provides through its finer temporal sampling.

In low light the two cameras reach similar quality, but we used a \texttt{cutoff\_hz} at the low end of the range discussed in Section~\ref{sec:cutoff}. The original DVS reached about $300$\,Hz at $1$\,lux~\cite{lichtsteiner_dvs_2008, gallego_survey_2022}, suggesting a better low-light temporal response than the high-resolution sensor model used here. However, its lower spatial resolution comes at a cost: it can limit the fine detail available to the SAI reconstruction. As shown by Gehrig and Scaramuzza~\cite{gehrig_high_res_2022}, lower-resolution event cameras can perform better in low light and at high speed for several tasks. For SAI, this benefit must therefore be weighed against the loss of spatial detail. Our comparison evaluates one choice within this tradeoff.

At approximately matched bandwidth budgets in the bright $1$\,m/s run, the event camera has no advantage in target-detail preservation. Especially at the lower budget, the $40$ frames retain slightly more detail than the $10$\,MEv/s event stream. At this speed, even the reduced frame rate supplies enough viewpoints to retain the bicycle, so the finer temporal sampling of the event camera provides less benefit. Its data rate also depends on the scene and, among our runs, is highest in the fast, bright condition where its clearest advantage appears. A system that adopts an event camera for high-speed SAI must therefore account for this high event rate and the event removal required by its link budget.

Another factor could be that at higher speeds the event camera records fewer events from the occluders, since motion parallax makes them move much faster across the sensor than the target. The limited photoreceptor bandwidth may attenuate these fast brightness changes more strongly. At lower speeds, more occluder events may survive and consume part of the ERC budget. This possible filtering occurs before the ERC, which drops events randomly in our model. But this is still an open question.

\begin{figure}[t]
\centering
\includegraphics[width=0.6\columnwidth]{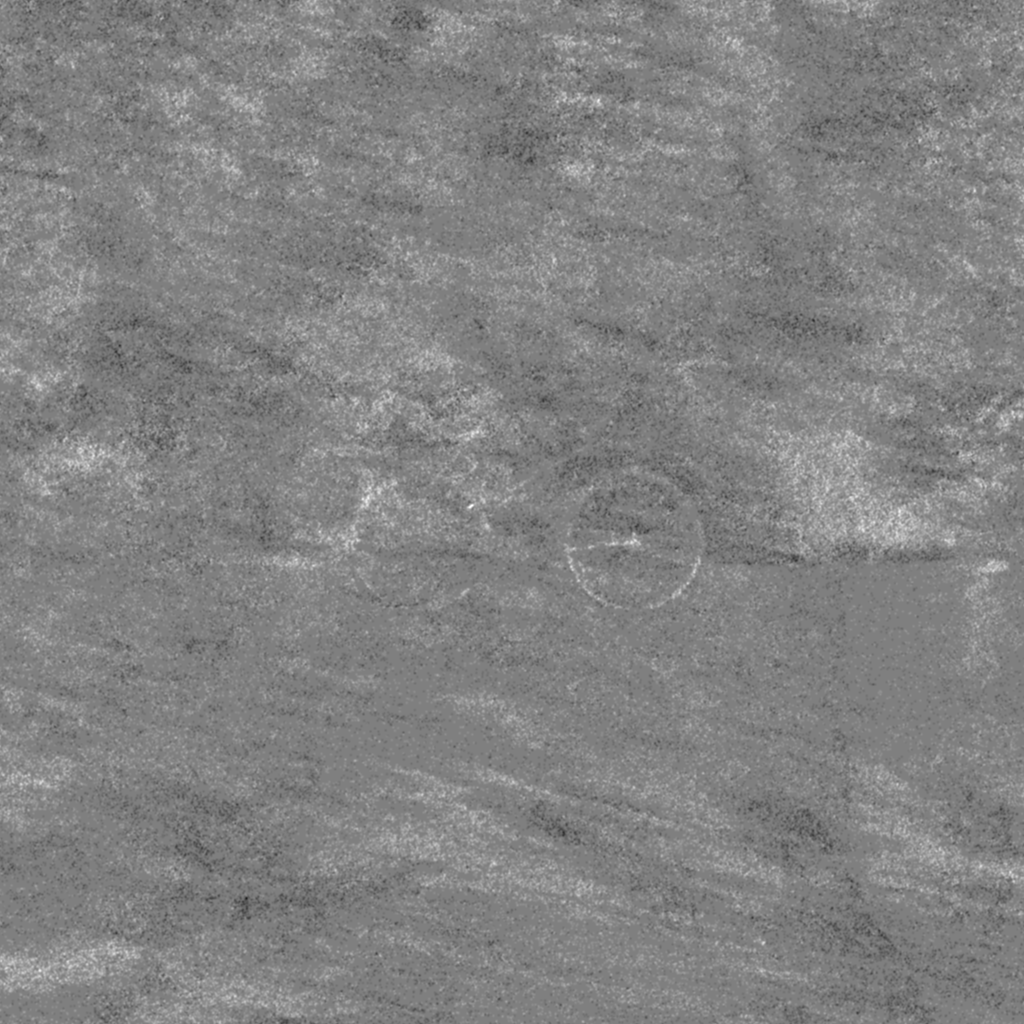}
\caption{Event SAI in the dark at high speed ($25$\,m/s, $5$\,lux, $100$\,Hz cutoff). Only a faint outline of one wheel remains, and the foliage is not averaged out.}
\Description{A square gray reconstruction dominated by streaked noise. A faint circle, the outline of one bicycle wheel, is visible slightly right of center.}
\label{fig:res-dark-fast}
\end{figure}

\paragraph{Dark at high speed.}
We tested high speed only in bright light. To see whether the event advantage survives in the dark, we repeated the $25$\,m/s run with the $5$\,lux settings of the night scenario, the $100$\,Hz cutoff and the $10$\,Hz per pixel shot noise, for the event camera only (Figure~\ref{fig:res-dark-fast}). The bicycle is barely recognizable: only a faint outline of one wheel remains, and the foliage is not averaged out. At $5$\,lux the time constant of the photoreceptor is about $1.6$\,ms, while a pixel crossing at $25$\,m/s takes about $0.7$\,ms, so most edges are filtered away before they trigger an event. The stream drops to about $23$\,MEv/s, so the link is not the limit here, the pixel is. The frame camera would record three noisy frames with $9$\,px of motion blur in this condition, so neither camera works, and the event advantage at high speed is limited to good light with this sensor model.

\paragraph{Practical impact.}
A speed of $25$\,m/s is relevant to fast-flying drones, and event cameras have already been flown on drones~\cite{sun_frames_events_2021}, though not for SAI. If at least ten frames are needed over the aperture to suppress the occluders, a $30$\,fps camera on our $0.5$\,m aperture is limited to $1.5$\,m/s, and the $160$\,fps camera we simulate to $8$\,m/s. The event camera gave a usable result at $25$\,m/s, a factor of about $17$ over the $30$\,fps camera. The faster traversal also matters when the target moves.

\paragraph{Pose estimation.}
Accurate pose estimation remains a practical limitation for high-speed event-based SAI. Recent event-inertial methods explicitly address aggressive motion, including continuous-time estimation of linear velocity~\cite{lu_highspeed_2025}, while newer benchmarks evaluate pose estimators on drone trajectories reaching $8.64$\,m/s~\cite{zhong_evslam_2026}. These studies support the feasibility of estimating motion from events and inertial measurements, but also show that reliable estimation under general high-speed motion remains challenging. Whether the resulting trajectories provide sufficient accuracy for SAI at the $25$\,m/s operating point tested here remains unverified.

\paragraph{Denser occlusion.}
Denser occlusion was not tested in our experiments. For frames, Kurmi et al.~\cite{kurmi_statistical_2019} show that denser occlusion lowers the visibility, and adding more frames does not improve the result after a certain threshold and never reaches the same result as with less occlusion. For events, the only study we know of is Yu et al.~\cite{yu_seethrough_2022}, whose reconstruction improved with denser occlusion. That test used a fence and cut cardboard with a trained network, so it does not tell us how plain integration behaves behind foliage.

\paragraph{Limitations.}
The comparisons are qualitative, and we only test one scene with mild occlusion. The estimation of the cutoff frequency is grounded but not verified with a real sensor. The ERC modeling is also an estimation. Therefore a difference would have to be substantial to infer an advantage, which we only had in the high-speed case. Also, the comparison excluded any HDR comparison, and HDR is definitely a big factor when considering event cameras.

\section{Conclusion}
We built a simulation pipeline in which a frame camera and an event camera see the same rendered scene and the same camera movement, and compared them for synthetic-aperture imaging. An event-native version of the AOS SAI was added. The comparison shows an advantage for the event camera at high-speed movements on a short path. This holds in good light. In the dark at the same speed, the event reconstruction also failed to recover the target clearly; the frame camera would remain undersampled. For the tested $0.5$\,m aperture and an assumed minimum of ten frames, this corresponds to approximately $17$ times the traversal speed of a $30$\,fps camera. The expected bandwidth efficiency of event cameras was not confirmed, because of the many events produced by the occlusion and the non-static camera setting. For low light no advantage could be found with the two selected sensors, although this does not mean that other event cameras with smaller resolution do not have an advantage at low light. In general the HDR advantage is still a very compelling argument to choose event cameras for synthetic aperture. If there is a usage scenario that needs the higher dynamic range, there is no reason to believe that this advantage should not transfer directly to SAI. In general event cameras are a viable option for SAI, but the advantages are not as big as expected. Future work should validate the simulation with real recordings, in particular the photoreceptor bandwidth of the event sensor in low light.

\bibliographystyle{ACM-Reference-Format}
\bibliography{references}

\end{document}